\documentclass{article}

\usepackage{PRIMEarxiv}

\usepackage[utf8]{inputenc} 
\usepackage[T1]{fontenc}    
\usepackage{hyperref}       
\usepackage{url}            
\usepackage{booktabs}       
\usepackage{amsfonts}       
\usepackage{nicefrac}       
\usepackage{microtype}      
\usepackage{lipsum}
\usepackage{fancyhdr}       
\usepackage{graphicx}       

\usepackage{amsmath}
\usepackage{cleveref}
\usepackage{graphicx} 
\usepackage{booktabs} 
\usepackage{authblk}
\usepackage{pifont}
\newcommand{\cmark}{\ding{51}}
\newcommand{\xmark}{\ding{55}}

\graphicspath{{media/}}     

\title{\textbf{SkyShield}: Occupancy as a Safety Interface for Low-Altitude UAV Autonomy
}

\author{
  Jie Gao\textsuperscript{1,$\dagger$}, Jie Ma\textsuperscript{1,$\dagger$}, Kaihui Lin\textsuperscript{1}, Kai Ye\textsuperscript{1}, Miaohui Zhang\textsuperscript{2}, Pingyang Dai\textsuperscript{1}, Liujuan Cao\textsuperscript{1} \\
  \textsuperscript{1}Xiamen University \quad \textsuperscript{2}Jiangxi Academy of Sciences \\
  \texttt{\{jiegao, jiema100, kaihuilin, yekai\}@stu.xmu.edu.cn} \\ 
  \texttt{jxaszmh@163.com} \quad 
  \texttt{\{pydai, caoliujuan\}@xmu.edu.cn} \\

}

\makeatletter
\def\blfootnote{\gdef\@thefnmark{}\@footnotetext}
\makeatletter

\begin{document}
\maketitle
\blfootnote{\textsuperscript{$\dagger$}Equal contribution.}

\begin{abstract}
For low-altitude Unmanned Aerial Vehicle (UAV) autonomy, 3D spatial understanding is not merely a perception objective, but the safety interface between human instructions and physical flight.
In human-scale urban airspace below 20 meters, thin geometry, occlusions, vegetation, and urban clutter define whether an aerial agent can safely enter the space ahead.
However, existing UAV datasets mainly provide 2D annotations or 3D boxes, while driving-oriented occupancy benchmarks assume stable ground-level sensor rigs.
Both miss the defining regime of low-altitude flight: a front-facing monocular camera observing occupied and free space from a moving aerial body with frame-wise changing 6-DoF pose and camera extrinsics.
To bridge this gap, we introduce \textbf{SkyShield}, to the best of our knowledge the first front-view monocular semantic occupancy benchmark for urban UAV flight below 20 meters.
Built on CARLA, SkyShield contains 36K front-view UAV samples across diverse urban scenes and weather conditions, pairing each image with frame-wise 6-DoF UAV pose, frame-wise dynamic camera geometry, UAV states, and front-frustum semantic occupancy labels.
We further propose \textbf{KAR-mIoU}, a UAV-centric and dynamics-aware metric that re-weights voxel-level evaluation by kinematic reachability and time-to-collision, revealing safety-critical risks hidden by conventional mIoU.
To tackle this challenging new setting, we provide \textbf{SkyOcc}, a geometry-first monocular baseline that integrates frame-wise UAV attitude into projection, fuses temporal occupancy features, and applies safety-prior optimization to preserve sparse collision-critical structures.
Together, SkyShield, KAR-mIoU, and SkyOcc establish occupancy as a safety interface for low-altitude aerial autonomy.
Code and dataset will be released publicly.
\end{abstract}


\section{Introduction}
Unmanned aerial vehicles (UAVs) are entering the airspace where people live and work.
Delivery drones move between buildings, inspection platforms fly close to facades, and many urban UAV missions take place below 20 meters of altitude~\cite{uav_survey,uav_perception,life_20m_v1,life_20m_v2}.
Recent advances in vision-language navigation and vision-language-action systems further expand what aerial agents are expected to do, enabling them to follow high-level instructions in open environments~\cite{uav_vln,uav_vla,robot_vln}.
Yet before a UAV acts, a more fundamental question must be answered: \textit{Does it understand the 3D space it is about to enter?}
For low-altitude UAV autonomy, dense spatial understanding is therefore not merely a perception objective, but the safety interface between human intent and physical flight.

\begin{figure*}[t!]
    \centering
    \includegraphics[width=\textwidth]{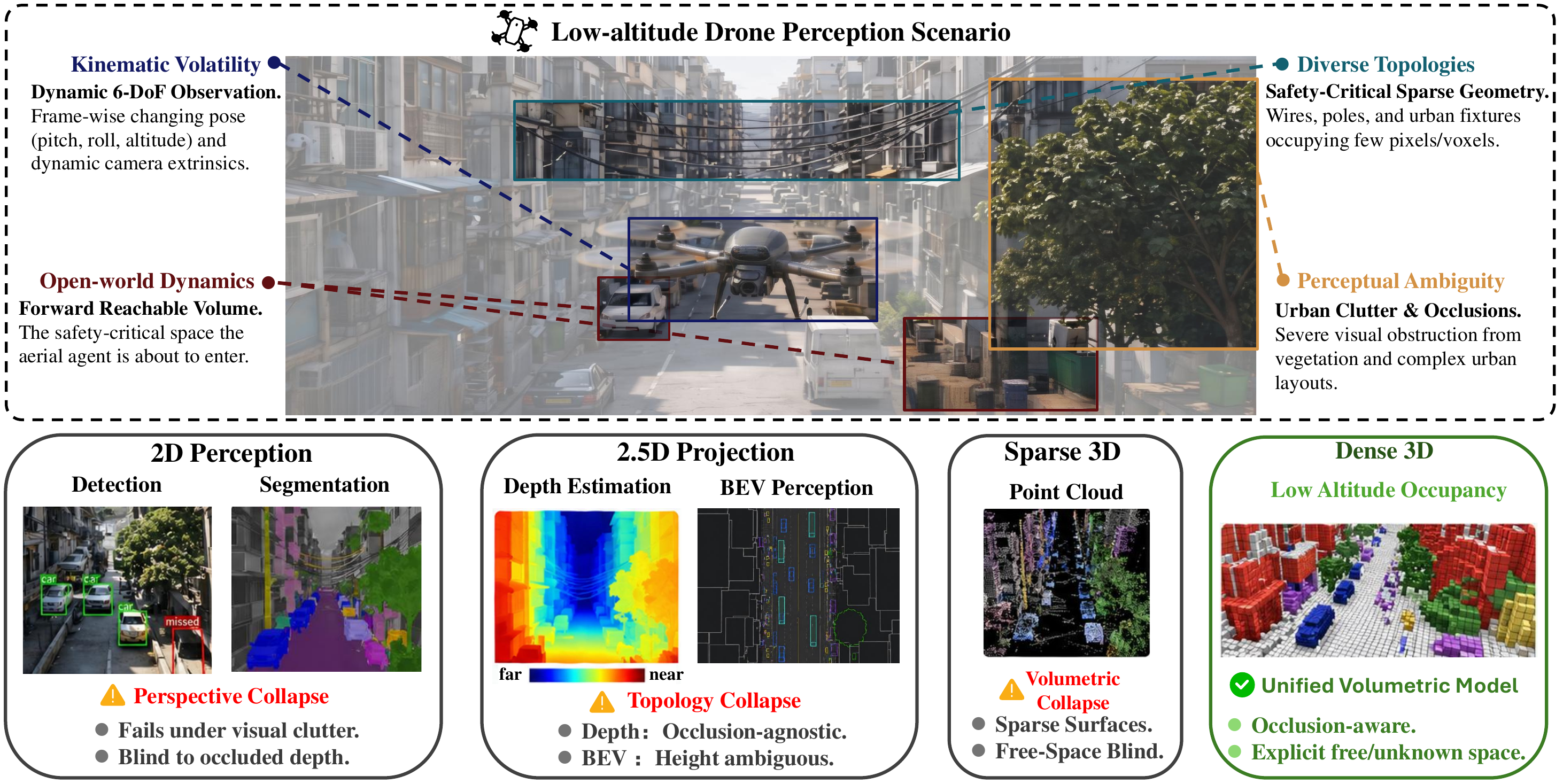}
    \caption{
    \textbf{Motivation.}
    Low-altitude UAV scenarios expose the limitations of traditional UAV perception and driving-oriented occupancy models.
    }
    \label{fig:motivation}
\end{figure*}

Existing UAV perception has achieved strong progress in detection, segmentation, depth estimation, and instance-level 3D understanding~\cite{uav_detection,uav_depth,ye_uav3d_2024}.
However, these tasks still describe the world mainly through objects, pixels, or surfaces.
As illustrated in \Cref{fig:motivation}, a low-altitude UAV needs a denser and more actionable representation: it must know which regions are occupied, which are free, and which parts of the forward space are unsafe to enter.
3D occupancy prediction is a natural representation for this purpose.
By assigning each voxel an occupancy state and semantic label, occupancy unifies geometry, free space, and obstacles into a form directly usable by downstream planning, and has become an important direction in autonomous driving perception~\cite{occupancy_network,sima_occnet_2023,wei_surroundocc_2023,tian_occ3d_2023}.

However, transferring occupancy from roads to low-altitude flight is not a simple change of viewpoint. Driving-oriented occupancy often assumes near-planar motion, stable sensor rigs, and structured road layouts \cite{cao_monoscene_2022, wei_surroundocc_2023}. A UAV violates these assumptions by design: its camera changes pitch, roll, altitude, and position continuously \cite{rosinol2020kimera}, while the safety-critical space is not a fixed ground-plane region but the forward volume the vehicle may reach in the next moments \cite{zhou2021egoplanner}.
Thus, the central difficulty is not only to recover 3D structure from a monocular image, but to align visual observations with the forward 3D space of a moving aerial observer.
A missed thin obstacle in the near flight corridor can be more important than a large object far outside the reachable region.
This makes low-altitude UAV occupancy a safety-centered spatial understanding problem rather than a direct extension of ground-vehicle occupancy.

This setting exposes three gaps.
First, existing benchmarks do not provide the supervision required for this problem.
UAV datasets mainly offer 2D annotations or 3D boxes, while driving occupancy datasets provide dense labels under ground-level motion and mostly stable extrinsics~\cite{uav_detection,ye_uav3d_2024}.
To the best of our knowledge, no public benchmark targets front-view monocular semantic occupancy prediction for urban UAV flight below 20 meters with frame-wise dynamic camera geometry.
Second, standard metrics are weakly aligned with flight safety.
Mean IoU treats all voxels equally, although errors near the UAV's reachable forward space are far more critical than errors in distant or unreachable regions.
Third, existing monocular occupancy models are not designed for the sparse structures that often dominate low-altitude risk.
Poles, wires, traffic lights, branches, and street fixtures occupy few pixels and voxels, appear under partial occlusion, and are easily weakened by pitch- and roll-induced feature misalignment.

To bridge these gaps, we introduce \textbf{SkyShield}, a high-fidelity benchmark built on CARLA for front-view monocular semantic occupancy prediction in urban low-altitude UAV flight.
SkyShield contains 36K front-view UAV samples across diverse urban scenes and weather conditions, pairing each image with frame-wise 6-DoF UAV pose, frame-wise dynamic camera geometry, UAV states, and front-frustum semantic occupancy labels.
We further propose \textbf{Kinematics-Aware Risk mIoU (KAR-mIoU)}, a UAV-centric and dynamics-aware metric that re-weights voxel-level evaluation by kinematic reachability and time-to-collision, revealing safety-critical errors hidden by conventional mIoU.
Finally, we provide \textbf{SkyOcc}, a reproducible geometry-first reference baseline for monocular UAV-centric semantic occupancy prediction, which instantiates the proposed task with attitude-aware projection, temporal occupancy fusion, and long-tail safety-aware supervision.
Together, SkyShield, KAR-mIoU, and SkyOcc define a new benchmark setting: occupancy as the safety interface for low-altitude UAV autonomy.

Our contributions are summarized as follows:
\begin{itemize}
    \item We introduce \textbf{SkyShield}, to the best of our knowledge the first front-view monocular semantic occupancy benchmark for urban UAV flight below 20 meters, comprising 36K samples across diverse scenes and weather conditions, each paired with frame-wise UAV states, dynamic camera geometry, and front-frustum semantic occupancy labels.

    \item We propose \textbf{KAR-mIoU}, a safety-aware metric that couples voxel-level occupancy evaluation with UAV reachability and time-to-collision risk.

    \item We provide \textbf{SkyOcc}, a reproducible monocular UAV-OCC reference baseline that makes dynamic observer geometry and long-tail safety-critical structures explicit in the learning pipeline.
\end{itemize}

\section{Related Work}
\subsection{UAV Perception Benchmarks}

UAV perception benchmarks have largely been defined by recognition tasks. Early datasets such as UAVDT~\cite{du_uavdt_2018}, VisDrone~\cite{zhu_visdrone_2021}, and DroneVehicle~\cite{sun_dronevehicle_2022} focus on 2D detection and tracking under aerial viewpoints, diverse scales, and complex imaging conditions. Recent benchmarks further extend this setting: MAVREC~\cite{dutta_mavrec_2024} introduces synchronized ground-and-aerial views, CoPerception-UAV~\cite{hu_coperceptionuav_2022} studies collaborative perception, and UAV3D~\cite{ye_uav3d_2024} provides 3D boxes in simulated urban scenes. These datasets have advanced aerial detection, tracking, and 3D localization, but they still describe the world mainly through pixels, instances, or boxes.

For low-altitude autonomy, this task definition is incomplete. A UAV does not only need to recognize known objects; it must know which part of the forward 3D space is occupied, free, reachable, or unsafe to enter~\cite{life_20m_v2}. This distinction is critical below 20 meters, where thin geometry, vegetation, urban clutter, and partial occlusions may define the boundary between safe flight and collision. Existing UAV benchmarks lack dense semantic occupancy supervision, while their standard metrics remain weakly coupled with flight safety. In contrast, \textbf{SkyShield} formulates UAV perception as front-view monocular semantic occupancy prediction, providing 36K samples with dynamic camera geometry, UAV states, and front-frustum occupancy labels. We further introduce \textbf{KAR-mIoU} to evaluate occupancy errors by UAV reachability and time-to-collision, making safety-critical failures visible beyond uniform voxel averaging.

\subsection{3D Spatial Perception and Occupancy}

3D spatial perception has progressed from object-centric recognition to more complete scene representations~\cite{spatial_nature, perceiving_nature}. Object-centric methods, such as 3D detection and tracking, localize predefined categories but cannot describe free space, non-object geometry, or unknown obstacles. Depth estimation and BEV representations~\cite{philion_lss_2020,li_bevformer_2022,li_bevdepth_2023,liu_bevfusion_2023,wang_detr3d_2022,liu_petr_2022} introduce geometric structure, yet depth lacks semantic volumetric reasoning and BEV compresses height, making it less suitable for overhangs, thin structures, and non-planar aerial scenes. Volumetric occupancy prediction addresses these limitations by assigning each voxel an occupancy state and semantic label. Representative works including MonoScene~\cite{cao_monoscene_2022}, TPVFormer~\cite{huang_tpvformer_2023}, SurroundOcc~\cite{wei_surroundocc_2023}, OpenOccupancy~\cite{wang_openoccupancy_2023}, Occ3D~\cite{tian_occ3d_2023}, OccNet~\cite{sima_occnet_2023}, VoxFormer~\cite{li_voxformer_2023}, and COTR~\cite{ma_cotr_2024} show that occupancy can unify geometry, semantics, obstacles, and free space in a planning-friendly representation.

However, most occupancy benchmarks and methods~\cite{huang_tpvformer_2023,wei_surroundocc_2023,li_voxformer_2023} are developed for ground vehicles, where motion is near-planar, sensor rigs are relatively stable, and scene structure is dominated by road layouts.
Low-altitude UAVs break these assumptions.
Their front cameras undergo continuous changes in pitch, roll, altitude, and position, while the safety-relevant space is the forward volume the vehicle may soon enter rather than a fixed ground-plane region.
This makes monocular UAV occupancy fundamentally observer-centered: the model must align visual observations with the moving UAV-centered 3D space while preserving sparse safety-critical geometry such as poles, wires, traffic lights, branches, and street fixtures.
In this setting, known rigid-body attitude changes should not be treated merely as auxiliary metadata: they provide direct geometric cues for stabilizing projection, while rare collision-critical structures require explicit safety priors to avoid being overwhelmed by dominant classes.
\textbf{SkyOcc} is designed as a reproducible geometry-first baseline for this setting, bringing front-view volumetric occupancy prediction to dynamic low-altitude UAV flight.

\section{SkyShield Benchmark}
\label{sec:uav_occ_benchmark}

\subsection{Task Definition}
\label{sec:task_definition}

\textbf{SkyShield} targets front-view monocular semantic occupancy prediction for low-altitude UAV flight.
Given a UAV front-view image $\mathcal{I}^{\mathrm{front}}_t$, the task is to infer occupied semantic voxels and observed free-space voxels inside the current front-camera frustum:
\begin{equation}
    \mathcal{O}^{\mathrm{front}}_t =
    \left\{
    (\mathbf{v}_i, y_i)
    \mid
    \mathbf{v}_i \in \mathbb{Z}^{3},
    \;
    y_i \in \mathcal{C} \cup \{c_{\mathrm{free}}\}
    \right\},
    \label{eq:uav_occ_task}
\end{equation}
where $\mathbf{v}_i$ denotes the voxel coordinate, $y_i$ is the voxel label, $\mathcal{C}$ is the occupied semantic class set, and $c_{\mathrm{free}}$ denotes observed free space.
Unknown regions are excluded from calibrated semantic supervision.
This front-frustum formulation is intentionally observer-centered: the benchmark evaluates whether a UAV can recover the 3D space it is about to enter from a single forward-looking camera.

\paragraph{Dynamic camera geometry.}
We use $\mathbf{T}^{\mathcal{A}\leftarrow\mathcal{B}}$ to denote a rigid transform from frame $\mathcal{B}$ to frame $\mathcal{A}$.
The front camera is rigidly mounted on the UAV, so the camera-to-body calibration $\mathbf{T}^{\mathcal{B}\leftarrow\mathcal{C}}$ is fixed.
The frame-wise dynamic quantity is the UAV body pose $\mathbf{T}^{\mathcal{W}\leftarrow\mathcal{B}}_t$.
Thus, the dynamic camera geometry is derived as
\begin{equation}
    \mathbf{T}^{\mathcal{W}\leftarrow\mathcal{C}}_t
    =
    \mathbf{T}^{\mathcal{W}\leftarrow\mathcal{B}}_t
    \mathbf{T}^{\mathcal{B}\leftarrow\mathcal{C}} .
    \label{eq:main_dynamic_camera_geometry}
\end{equation}
Roll, pitch, and yaw enter the projection through this transformation chain exactly once.
Full coordinate definitions are provided in \Cref{app:coordinate_convention}.

\begin{table}[t]
\centering
\small
\caption{
\textbf{Dataset-level comparison.}
SkyShield is the only benchmark in this comparison that jointly provides UAV-view imagery, 3D annotation, dense occupancy labels, and frame-wise dynamic camera geometry.
Diversity measures scene richness, including variations in urban layout, weather, altitude, viewpoint, clutter, and safety-critical structures.
}
\label{tab:dataset_comparison}
\resizebox{\linewidth}{!}{
\begin{tabular}{lcccccccc}
\toprule
Dataset & UAV View & 3D Anno. & Occ. Label & Dyn. Cam. Geom. & Diversity & Train & Val & Test \\
\midrule
UAVDT~\cite{du_uavdt_2018}              & \cmark & \xmark & \xmark & \xmark & 1  & 24,143 & 0     & 53,676 \\
VisDrone~\cite{zhu_visdrone_2021}       & \cmark & \xmark & \xmark & \xmark & 3  & 6,471  & 548   & 1,610  \\
UAV3D~\cite{ye_uav3d_2024}              & \cmark & \cmark & \xmark & \cmark & 2  & 14,000 & 3,000 & 3,000  \\
KITTI~\cite{geiger_kitti_2012}          & \xmark & \cmark & \xmark & \xmark & 1  & 7,481  & 0     & 7,518  \\
nuScenes~\cite{caesar_nuscenes_2020}    & \xmark & \cmark & \xmark & \xmark & 6  & 28,130 & 6,019 & 6,008  \\
\textbf{SkyShield} (Ours)               & \textbf{\cmark} & \textbf{\cmark} & \textbf{\cmark} & \textbf{\cmark} & \textbf{10} & \textbf{29,000} & \textbf{3,200} & \textbf{3,800} \\
\bottomrule
\end{tabular}
}
\end{table}

\Cref{tab:dataset_comparison} exposes the benchmark gap directly.
Existing UAV datasets provide the aerial viewpoint but usually stop at 2D or 3D boxes, while existing occupancy datasets provide dense volumetric labels for ground vehicles with static sensor rigs.
SkyShield fills this missing regime with front-view monocular UAV occupancy, frame-wise dynamic camera geometry, dense semantic occupancy supervision, rich scene coverage, and complete train/validation/test splits.

\subsection{Benchmark Construction}
\label{sec:benchmark_construction}

SkyShield is built on CARLA~\cite{dosovitskiy_carla_2017} for urban low-altitude UAV flight below $20$ meters.
It contains $36$K front-view UAV samples collected at $25$Hz from $8$ urban scenes, $10$ weather conditions, and $15$ clips per scene, with $29$K/$3.2$K/$3.8$K train/validation/test splits.
Each sample provides a monocular front-view RGB image, frame-wise 6-DoF UAV body pose, fixed front-camera calibration, derived dynamic camera geometry, UAV motion states, and front-frustum semantic occupancy labels.

Data collection is centered on safety-critical low-altitude structure.
Open roads alone do not characterize UAV risk: facades, vegetation, poles, traffic lights, wires, branches, and cluttered corridors often decide whether the forward space is traversable.
To obtain reliable occupancy supervision while preserving a monocular inference setting, SkyShield mounts two nearly blind-spot-free LiDAR sensors on the UAV, one above and one below the platform.
These LiDAR streams and simulator semantics are used only for annotation; the learning input remains a single monocular front-view RGB image.

\begin{figure*}[t]
    \centering
    \includegraphics[width=\textwidth]{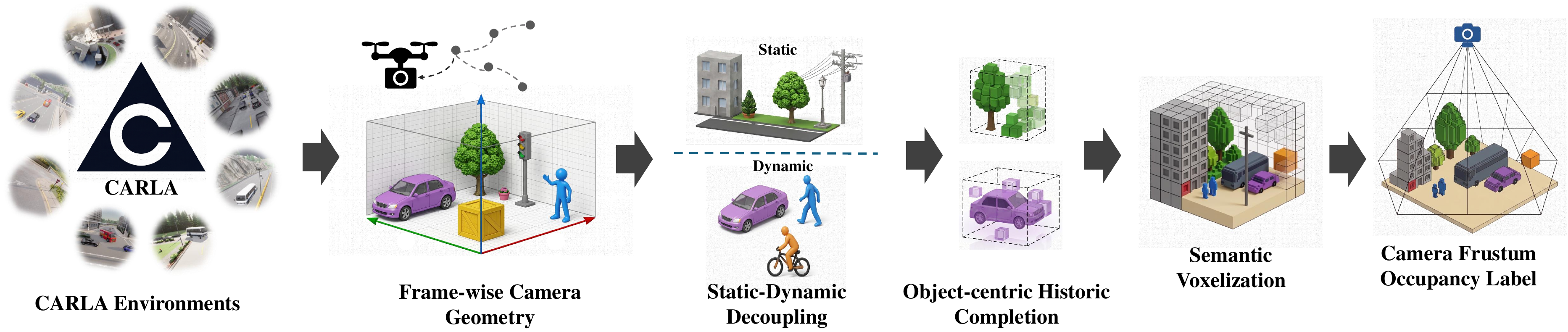}
    \caption{
    \textbf{SkyShield benchmark construction.}
    SkyShield converts low-altitude UAV simulation into front-frustum semantic occupancy supervision through fixed camera calibration, frame-wise UAV pose, static-dynamic decoupling, semantic voxelization, free-space ray casting, and front-frustum filtering.
    Dual upper/lower LiDAR and simulator semantics are used only for annotation; the learning input remains monocular front-view RGB.
    }
    \label{fig:benchmark_pipeline}
\end{figure*}

Occupancy labels are generated through a coordinate-consistent pipeline.
All annotations follow a single transformation chain from the world frame to the UAV-centered occupancy frame and then to the front-camera frame.
World-space semantic geometry is first transformed into $\mathcal{G}_t$ for voxelization, and the resulting voxel centers are projected into $\mathcal{C}_t$ for frustum filtering and visibility validation.
Static infrastructure and dynamic actors are decoupled to reduce temporal ghosting and hollow-object artifacts.
The fused scene is voxelized, assigned semantic labels, enriched with ray-casted observed free space, and filtered by the front-camera frustum.
The final output is front-frustum occupied/free semantic supervision aligned with the monocular front image, as shown in \Cref{fig:benchmark_pipeline}.

\subsection{Safety-Aware Evaluation}
\label{sec:safety_aware_evaluation}

Conventional mean Intersection-over-Union (mIoU) treats every valid voxel equally, which is poorly aligned with low-altitude UAV safety.
A false negative near the UAV may leave little reaction time, while a similar error far outside the reachable region is less urgent.
SkyShield therefore reports both standard mIoU and \textbf{Kinematics-Aware Risk mIoU (KAR-mIoU)}, which penalizes false positives and false negatives according to time-to-collision (TTC).

SkyShield adopts an \textbf{Omnidirectional Worst-Case TTC (OWC-TTC)} to reflect open low-altitude airspace, where side drift, hovering instability, and non-cooperative aerial objects can make purely forward TTC insufficient.
For voxel $i$, let $\mathbf{x}_i$ be its center in the UAV-centered frame and $\mathbf{v}_t$ be the UAV velocity.
We define the OWC-TTC and penalty as
\begin{equation}
    \mathrm{TTC}^{\mathrm{owc}}_i =
    \frac{\|\mathbf{x}_i\|_2}{\max(\|\mathbf{v}_t\|_2,\epsilon)},
    \qquad
    p_i =
    1 +
    \mathbb{I}
    \left[
    \mathrm{TTC}^{\mathrm{owc}}_i \leq T_h
    \right]
    \gamma
    \exp
    \left(
    -\lambda \mathrm{TTC}^{\mathrm{owc}}_i
    \right),
    \label{eq:owc_ttc_penalty}
\end{equation}
where $T_h$ is the safety horizon, $\gamma$ controls penalty scale, and $\lambda$ controls temporal decay.
We set $T_h=5.0$s following established TTC-based safety metrics for short-horizon collision warning and reactive avoidance~\cite{tottrup2022ttc}.
For semantic class $c$, KAR-IoU is defined as
\begin{equation}
    \mathrm{KAR\text{-}IoU}_c =
    \frac{
        \mathrm{TP}_c
    }{
        \mathrm{TP}_c
        +
        \sum_{i \in \mathrm{FP}_c} p_i
        +
        \sum_{i \in \mathrm{FN}_c} p_i
    }.
    \label{eq:kar_iou}
\end{equation}
Correct predictions are not rewarded by proximity; only mistakes are penalized.
The final KAR-mIoU averages KAR-IoU over evaluated occupied semantic classes.
Unknown voxels are ignored, and free space is reported separately but excluded from the KAR-mIoU mean to avoid domination by the overwhelmingly frequent free-space class.
Thus, KAR-mIoU asks a sharper safety question: when occupancy prediction fails, does it fail where the UAV has the least time to recover?

\section{Method}
\label{sec:method}

\subsection{Overview}
\label{sec:method_overview}

SkyOcc is designed for front-view monocular semantic occupancy prediction in dynamic low-altitude UAV flight.
Unlike driving-oriented occupancy, where sensor rigs are relatively stable, UAV cameras undergo continuous roll, pitch, altitude, and position changes.
These changes directly affect the projection between the monocular image and the UAV-centered 3D space.
SkyOcc follows a \textbf{geometry-aware reference design}: the frame-wise UAV pose and fixed front-camera calibration define a consistent projection chain between the monocular image and the UAV-centered occupancy frame.
This does not introduce an additional pose correction outside the camera model; instead, UAV attitude enters the image-to-volume mapping exactly once through the transformation chain.

\begin{figure*}[t]
    \centering
    \includegraphics[width=\textwidth]{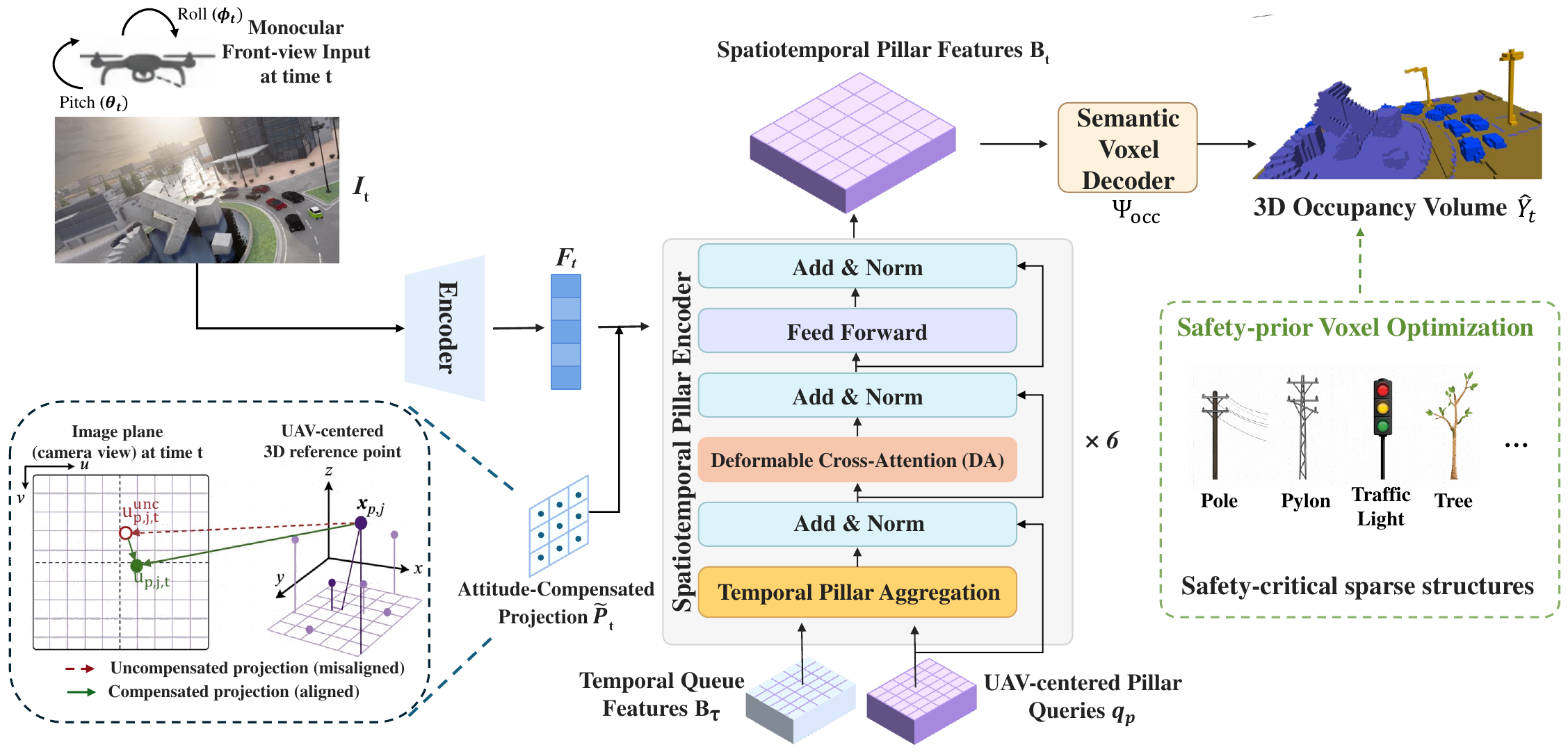}
    \caption{
    \textbf{SkyOcc pipeline.}
    SkyOcc uses attitude-aware projection to map UAV-centered pillar queries into the monocular front image, aggregates ego-motion-aligned historical pillar features, and optimizes semantic occupancy for safety-critical sparse structures.
    }
    \label{fig:skyocc_pipeline}
\end{figure*}

Given the current front-view image $I^{\mathrm{front}}_t$ and a cached queue of ego-motion-aligned historical pillar features, SkyOcc predicts the semantic occupancy volume
$\hat{\mathbf{Y}}_t \in \mathbb{R}^{H \times W \times Z \times |\mathcal{C}|}$
in the UAV-centered forward space, where $\mathcal{C}$ contains occupied semantic categories and observed free space.
The model consists of three core components:
\textbf{attitude-aware spatial projection},
\textbf{spatiotemporal pillar encoder},
and \textbf{safety-prior voxel optimization}, as shown in Figure~\ref{fig:skyocc_pipeline}.

\subsection{SkyOcc: Geometry-First Safety-Prior Occupancy}
\label{sec:skyocc}

\noindent\textbf{Attitude-aware spatial projection.}
Low-altitude UAV attitude changes the mapping between the UAV-centered occupancy frame and the front-view image.
SkyOcc does not introduce any post-hoc pose correction outside the camera model.
Instead, the frame-wise UAV pose and fixed camera calibration define a single transformation from the UAV-centered occupancy frame $\mathcal{G}_t$ to the front-camera frame $\mathcal{C}_t$, through which roll, pitch, and yaw enter the projection exactly once.

Let $\mathbf{x}^{\mathcal{G}}_{p,j}$ denote the $j$-th height reference point of pillar query $p$ in the UAV-centered occupancy frame $\mathcal{G}_t$.
Its image location is computed by
\begin{equation}
    \mathbf{u}_{p,j,t}
    =
    \pi
    \left(
    \mathbf{K}_t
    \begin{bmatrix}
    \mathbf{I}_3 & \mathbf{0}
    \end{bmatrix}
    \mathbf{T}^{\mathcal{C}\leftarrow\mathcal{G}}_t
    \bar{\mathbf{x}}^{\mathcal{G}}_{p,j}
    \right),
    \label{eq:attitude_aware_projection}
\end{equation}
where $\mathbf{K}_t \in \mathbb{R}^{3\times3}$ is the front-camera intrinsic matrix,
$\bar{\mathbf{x}}^{\mathcal{G}}_{p,j}=[(\mathbf{x}^{\mathcal{G}}_{p,j})^\top,1]^\top$ is the homogeneous coordinate,
$\mathbf{T}^{\mathcal{C}\leftarrow\mathcal{G}}_t \in SE(3)$ maps points from the UAV-centered occupancy frame to the front-camera frame, and $\pi(\cdot)$ denotes perspective normalization.
This projection provides a geometrically consistent interface between monocular image features and the UAV-centered forward occupancy volume.

\noindent\textbf{Spatiotemporal pillar encoder.}
For each frame, an image encoder extracts front-view features
\begin{equation}
    \mathbf{F}_t =
    \Phi_{\mathrm{img}}
    \left(
        I^{\mathrm{front}}_t
    \right).
\end{equation}
Queries $\mathbf{q}_p$ representing UAV-centered spatial pillars first aggregate temporal information from historical pillar features spatially aligned to the current UAV-centered coordinate system via ego-motion:
\begin{equation}
    \bar{\mathbf{q}}_{p,t}
    =
    \operatorname{TAttn}
    \left(
        \mathbf{q}_p,
        \left\{
        \widetilde{\mathbf{B}}_{\tau \rightarrow t}
        \right\}_{\tau=t-L+1}^{t-1}
    \right),
    \label{eq:temporal_pillar_attention}
\end{equation}
where $L$ denotes the temporal window length,
$\operatorname{TAttn}(\cdot)$ denotes temporal pillar aggregation and
$\widetilde{\mathbf{B}}_{\tau \rightarrow t}$ denotes historical pillar features aligned from frame $\tau$ to the current frame $t$.
Each pillar query $p$ contains height-wise reference points indexed by $j$.
Given the attitude-aware projection locations, the temporally aggregated query gathers current-frame image evidence through deformable cross-attention:
\begin{equation}
    \mathbf{B}_{p,t}
    =
    \frac{1}{|\mathcal{J}_{p,t}|}
    \sum_{j \in \mathcal{J}_{p,t}}
    \operatorname{DA}
    \left(
        \bar{\mathbf{q}}_{p,t},
        \mathbf{u}_{p,j,t},
        \mathbf{F}_t
    \right),
    \label{eq:front_view_lifting}
\end{equation}
where $\mathcal{J}_{p,t}$ is the set of valid projected height references and $\operatorname{DA}(\cdot)$ denotes deformable attention.
The resulting pillar features are stacked as $\mathbf{B}_t=\{\mathbf{B}_{p,t}\}_{p}$ and decoded into dense semantic occupancy:
\begin{equation}
    \hat{\mathbf{Y}}_t
    =
    \Psi_{\mathrm{occ}}
    \left(
        \mathbf{B}_t
    \right),
    \label{eq:voxel_decoding}
\end{equation}
where $\Psi_{\mathrm{occ}}(\cdot)$ acts as a semantic voxel decoder that expands the encoded pillar features into a full 3D occupancy volume.
With the image-to-volume mapping defined by attitude-aware projection and historical features aligned by ego-motion, the encoder mainly focuses on structural semantics and temporal consistency before voxel decoding.

\noindent\textbf{Safety-prior voxel optimization.}
Low-altitude UAV risk is often dominated by sparse structures such as poles, traffic signs, traffic lights, wires, branches, and other thin urban fixtures.
These classes occupy few voxels and are easily overwhelmed by ground, buildings, vegetation, and free space.
SkyOcc therefore uses static safety-prior class weights together with Lov\'asz-Softmax loss~\cite{lovasz}:
\begin{equation}
    \mathcal{L}_{\mathrm{occ}}
    =
    \frac{1}{|\Omega|}
    \sum_{i \in \Omega}
    \alpha_{y_i}
    \operatorname{CE}
    \left(
        \hat{\mathbf{y}}_i,
        y_i
    \right)
    +
    \lambda_{\mathrm{lov}}
    \mathcal{L}_{\mathrm{lov}}
    \left(
        \hat{\mathbf{Y}}_t,
        \mathbf{Y}_t
    \right),
    \label{eq:occ_loss}
\end{equation}
where $\Omega=\{i\mid y_i\neq c_{\mathrm{ignore}}\}$ is the supervised voxel set, $\alpha_{y_i}$ is a manually specified safety-prior weight, and $\mathcal{L}_{\mathrm{lov}}$ is an IoU-oriented surrogate.
The class weights increase the learning pressure on sparse collision-critical categories, while Lov\'asz-Softmax improves long-tail IoU optimization.
Together, attitude-aware projection and safety-prior optimization form a simple but effective reference baseline for low-altitude UAV occupancy.

\section{Experiments}
\label{sec:experiments}

\subsection{Experimental Setup}
\label{sec:experimental_setup}

\textbf{Dataset and protocol.}
We evaluate all models on the SkyShield test split for front-view monocular semantic occupancy prediction in urban low-altitude UAV flight.
The test set contains 3.8K frames and is disjoint from the 29K training set and 3.2K validation set.
Each sample contains a monocular front-view RGB image, frame-wise UAV pose, frame-wise dynamic camera geometry, UAV states, and front-frustum semantic occupancy labels.
The semantic label space contains nine evaluated categories:
\textit{building}, \textit{ground}, \textit{vegetation}, \textit{thin structures}, \textit{large vehicle}, \textit{car}, \textit{vulnerable road user (VRU)}, \textit{general}, and \textit{free space}.
The \textit{unknown} class is excluded from training and evaluation.

\textbf{Compared models.} We use a temporal monocular occupancy model as the baseline for all comparisons. The baseline is implemented by taking a BEVFormer backbone and attaching a semantic occupancy prediction head, using the same front-view input, temporal queue, image backbone, FPN neck, BEVFormer-style encoder, and occupancy head as SkyOcc, but without the proposed geometry-first and safety-prior designs. We then progressively add attitude-aware projection, static safety-prior class weights, and Lov\'asz-Softmax loss. The final configuration is denoted as \textbf{SkyOcc}.

\textbf{Implementation details.}
All models are trained on 8 NVIDIA A800 GPUs.
The per-GPU batch size is 8, resulting in a total batch size of 64.
Each model is trained for 24 epochs, and the full training takes approximately 6 hours.
We use ResNet50 as the image backbone, an FPN with four feature levels, a temporal pillar encoder, and an occupancy head for 3D semantic prediction.
We set the temporal window length to $L=4$.
The voxel grid size is $100 \times 100 \times 60$ with 0.5m resolution, and the point-cloud range is $[0,-25,-20,50,25,10]$ meters.
We optimize all models with AdamW using an initial learning rate of $2 \times 10^{-4}$ and a weight decay of $1 \times 10^{-2}$.
The learning rate follows a cosine annealing schedule.

\textbf{Metrics.}
We report class-wise IoU, mIoU, and KAR-mIoU.
mIoU is averaged over the nine evaluated classes including \textit{free space}.
KAR-mIoU follows the penalized mode introduced in \Cref{sec:safety_aware_evaluation}: true positives are not rewarded by proximity, while false positives and false negatives are penalized according to omnidirectional worst-case TTC.
Following established TTC-based safety metrics, we set the TTC horizon to 5.0s, corresponding to a short-horizon warning window for tactical collision risk assessment and reactive avoidance~\cite{tottrup2022ttc}.
In our test setting, we use $\lambda=0.5$ and penalty scale $\gamma=2.0$.
Following the benchmark protocol, \textit{free space} and \textit{unknown} are excluded from the KAR-mIoU mean, while occupied-class false positives and false negatives are still preserved in class-wise denominators.

\subsection{Quantitative Results}
\label{sec:quantitative_results}

\noindent\textbf{Main results.}
\cref{tab:main_results} reports the main results on the SkyShield test set. SkyShield improves the multi-frame baseline from 25.08 to 28.91 mIoU and from 15.04 to 19.10 KAR-mIoU. The gains are concentrated on sparse and safety-relevant categories: \textit{thin structures} improves from 5.57 to 12.80, \textit{car} from 13.19 to 23.24, and \textit{VRU} from 2.95 to 12.09. Meanwhile, \textit{free space} remains stable around 98.3 IoU, showing that the improvement does not come from sacrificing free-space reasoning.

\begin{table*}[t]
\centering
\small
\caption{
\textbf{Main results on the SkyShield test set.}
All numbers are percentages. mIoU is averaged over nine evaluated classes including \textit{free space}; KAR-mIoU excludes \textit{free space} and \textit{unknown} from the mean and penalizes FP/FN in short-TTC regions.
}
\label{tab:main_results}
\resizebox{\textwidth}{!}{
\begin{tabular}{lccccccccccc}
\toprule
Method & Building & Ground & Vegetation & Thin Struct. & Large Veh. & Car & VRU & General & Free Space & mIoU & KAR-mIoU \\
\midrule
Baseline
& 17.61 & \textbf{38.36} & 15.17 & 5.57 & 10.16 & 13.19 & 2.95 & 24.23 & \textbf{98.49} & 25.08 & 15.04 \\
+ Attitude-aware projection
& 18.43 & 37.72 & 15.71 & 7.17 & 11.39 & 15.32 & 9.54 & \textbf{25.25} & 98.48 & 26.56 & 16.59 \\
+ Safety-prior class weights
& 17.29 & 37.67 & 17.24 & 11.85 & 14.66 & 20.90 & 11.12 & 23.28 & 98.30 & 28.04 & 18.17 \\
\textbf{SkyShield}
& \textbf{19.11} & 36.74 & \textbf{18.69} & \textbf{12.80} & \textbf{14.82} & \textbf{23.24} & \textbf{12.09} & 24.42 & 98.30 & \textbf{28.91} & \textbf{19.10} \\
\bottomrule
\end{tabular}
}
\end{table*}

The results reveal a clear geometry-first pattern. Using attitude-aware projection improves both mIoU and KAR-mIoU, with especially large gains on geometry-sensitive foreground classes such as \textit{VRU}, \textit{car}, and \textit{thin structures}. This supports the core idea that measured UAV pose should enter the image-to-3D projection chain explicitly, rather than being treated only as auxiliary metadata. Safety-prior class weights further shift optimization toward rare collision-critical categories, producing strong gains on \textit{thin structures}, \textit{car}, \textit{VRU}, and \textit{large vehicle}. Lov\'asz-Softmax provides the final improvement by optimizing an IoU-oriented surrogate, which is better aligned with sparse occupancy than voxel-wise accuracy alone.

The gap between mIoU and KAR-mIoU should be interpreted under their different averaging protocols. mIoU includes \textit{free space}, whose IoU is above 98\% and can dominate the mean. KAR-mIoU excludes \textit{free space} and focuses on occupied categories, while further penalizing FP/FN in short-TTC regions. Under the adopted OWC-TTC penalty, the improvement from 15.04 to 19.10 indicates that SkyShield improves occupied-space reliability under the safety-aware protocol, not only the globally averaged reconstruction quality.

\noindent\textbf{Ablation study.}
\cref{tab:ablation} summarizes the contribution of each effective component. Attitude-aware projection improves the baseline by 1.48 mIoU and 1.55 KAR-mIoU. Static safety-prior class weights provide the largest long-tail gain, improving \textit{thin structures} by 6.28 points and \textit{VRU} by 8.17 points over the baseline. Lov\'asz-Softmax further improves the final model by 0.87 mIoU and 0.93 KAR-mIoU over the class-weighted variant.

\begin{table*}[t]
\centering
\small
\caption{
\textbf{Ablation study.}
Gains are measured relative to the baseline. The effective path is attitude-aware projection, static safety-prior class weights, and Lov\'asz-Softmax loss.
}
\label{tab:ablation}
\resizebox{\textwidth}{!}{
\begin{tabular}{lccccccccc}
\toprule
Configuration & Att. Proj. & Class Weight & Lov\'asz & mIoU & KAR-mIoU & $\Delta$ mIoU & $\Delta$ KAR & $\Delta$ Thin & $\Delta$ VRU \\
\midrule
Baseline
& \xmark & \xmark & \xmark
& 25.08 & 15.04 & 0.00 & 0.00 & 0.00 & 0.00 \\
+ Attitude-aware projection
& \cmark & \xmark & \xmark
& 26.56 & 16.59 & +1.48 & +1.55 & +1.60 & +6.59 \\
+ Safety-prior class weights
& \cmark & \cmark & \xmark
& 28.04 & 18.17 & +2.96 & +3.13 & +6.28 & +8.17 \\
\textbf{SkyShield}
& \cmark & \cmark & \cmark
& \textbf{28.91} & \textbf{19.10} & \textbf{+3.83} & \textbf{+4.06} & \textbf{+7.23} & \textbf{+9.14} \\
\bottomrule
\end{tabular}
}
\end{table*}

The ablation gives the main empirical insight of SkyShield. Roll and pitch are known rigid-body measurements, and using them directly in projection offers a simple and effective way to reduce attitude-induced misalignment. Once the geometry is stabilized, manual safety-prior weights become highly effective because they force sparse collision-critical categories to contribute meaningful gradients. Lov\'asz-Softmax complements this effect by optimizing an IoU-oriented objective. Together, these components convert low-altitude UAV occupancy from average scene reconstruction into safety-prior spatial understanding. This behavior is especially visible on long-tail safety structures: compared with the baseline, SkyShield improves \textit{thin structures}, \textit{large vehicle}, \textit{car}, and \textit{VRU} by 7.23, 4.66, 10.05, and 9.14 points, respectively. These categories occupy few voxels but often define the boundary of safe low-altitude flight.

\begin{figure*}[t]
    \centering
    \includegraphics[width=0.95\textwidth]{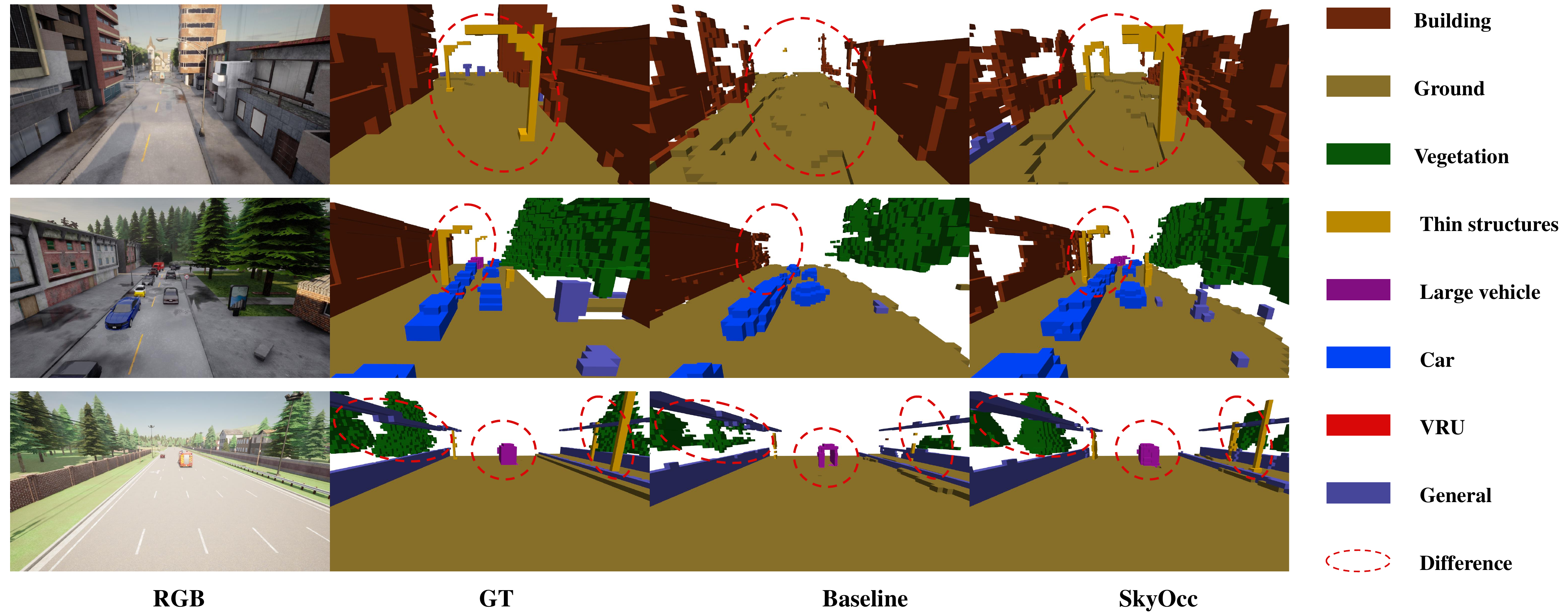}
    \caption{
    \textbf{Qualitative results on SkyShield.}
    From left to right, each group shows the front-view RGB image, ground truth, baseline occupancy prediction and SkyOcc prediction.
    The visualization highlights near-field obstacles, thin structures, occlusion boundaries, semantic occupancy, and free space in the UAV-centered front frustum.
    }
    \label{fig:qualitative}
    \vspace{-1em}
\end{figure*}

\subsection{Qualitative Results}
\label{sec:qualitative_results}

We present representative qualitative results on SkyShield in \Cref{fig:qualitative}.
The visualization compares the front-view image, baseline prediction, SkyOcc prediction, and ground-truth occupancy.
We include scenes with aggressive roll/pitch changes, cluttered urban corridors, vegetation occlusion, nearby vehicles or vulnerable road users, and thin safety-critical structures.
These cases reflect the core difficulty of low-altitude UAV flight: the model must infer not only large visible surfaces, but also the fine-grained forward space that constrains safe motion.

Qualitatively, SkyOcc produces a more stable front-frustum occupancy field under dynamic UAV viewpoints.
Attitude-aware projection reduces pose-induced projection drift, while safety-prior optimization better preserves thin structures and nearby foreground objects.
These visual results complement the quantitative findings: SkyOcc improves not only the average voxel score, but also the safety relevance of the predicted 3D space.
Together, the main results, ablations, and visualizations show that SkyShield exposes the right failure modes, KAR-mIoU measures potential risk in occupied space, and SkyOcc provides a strong reproducible baseline for low-altitude UAV occupancy.

\section{Conclusion}
\label{sec:conclusion}

We presented \textbf{SkyShield}, a front-view monocular semantic occupancy benchmark for urban UAV flight below 20\,m.
SkyShield targets the core condition of low-altitude autonomy: a UAV must infer the occupied, free, and safety-critical space ahead while its camera geometry changes with pitch, roll, altitude, and velocity.
To evaluate this setting, we introduced \textbf{KAR-mIoU}, a TTC-aware metric that complements conventional mIoU by emphasizing errors within the UAV's reachable and time-critical regions.
We further developed \textbf{SkyOcc}, a geometry-aware UAV-OCC reference baseline that uses attitude-aware front-view projection, temporal occupancy fusion, and long-tail safety-aware supervision to preserve sparse hazards such as wires, poles, traffic lights, and branches.
These contributions move UAV perception beyond recognizing objects in images toward understanding the 3D space an aerial agent is about to enter.


\bibliographystyle{unsrt}  
\bibliography{references}  

\newpage

\appendix

\section{Technical appendices and supplementary material}

\label{sec:supplementary}

This supplement provides the technical details that support the main paper without repeating its motivation. We focus on four aspects that are essential for judging the benchmark and the baseline: how SkyShield constructs physically plausible low-altitude UAV trajectories, how reliable front-view semantic occupancy labels are generated, how KAR-mIoU is implemented as a safety-aware evaluation metric, and how SkyOcc instantiates a reproducible geometry-first monocular occupancy baseline.

\subsection{SkyShield Dataset and Generation Details}
\label{app:skyshield_details}

SkyShield is generated in \textbf{CARLA} for urban low-altitude UAV flight below 20m. The dataset contains 36K front-view samples captured at 25Hz from 8 urban scenes, 10 weather conditions, and 15 clips per scene. Each clip lasts 12 seconds and contains 300 frames. The train/validation/test split contains 29K / 3.2K / 3.8K frames, respectively. The benchmark input is always monocular front-view RGB; auxiliary simulator signals and dual LiDAR streams are used only for annotation. Detailed dataset statistics are summarized in \Cref{tab:app_skyshield_statistics}.

\begin{table}[h]
\centering
\small
\caption{\textbf{SkyShield statistics.} Auxiliary sensors are used only for annotation; the learning input remains monocular front-view RGB.}
\label{tab:app_skyshield_statistics}
\begin{tabular}{lc}
\toprule
Item & Value \\
\midrule
Simulator & CARLA \\
Scenes & 8 \\
Weather conditions & 10 \\
Clips per scene & 15 \\
Clip duration & 12s \\
Capture frequency & 25Hz \\
Total samples & 36K \\
Train / Val / Test & 29K / 3.2K / 3.8K \\
Altitude range & $<20$m \\
Learning input & Monocular front-view RGB \\
Annotation signals & Dual LiDAR + simulator semantics \\
\bottomrule
\end{tabular}
\end{table}

The split is designed to evaluate both temporal generalization and unseen-trajectory robustness. Four clips are held out as unseen evaluation clips: one for validation and three for testing. The remaining 116 clips are split temporally, using the first 10s for training, 10--11s for validation, and 11--12s for testing. This protocol prevents frame overlap across splits and avoids the leakage that would arise from random sampling of near-duplicate adjacent frames.

\begin{figure}[h]
    \centering
    \includegraphics[width=\textwidth]{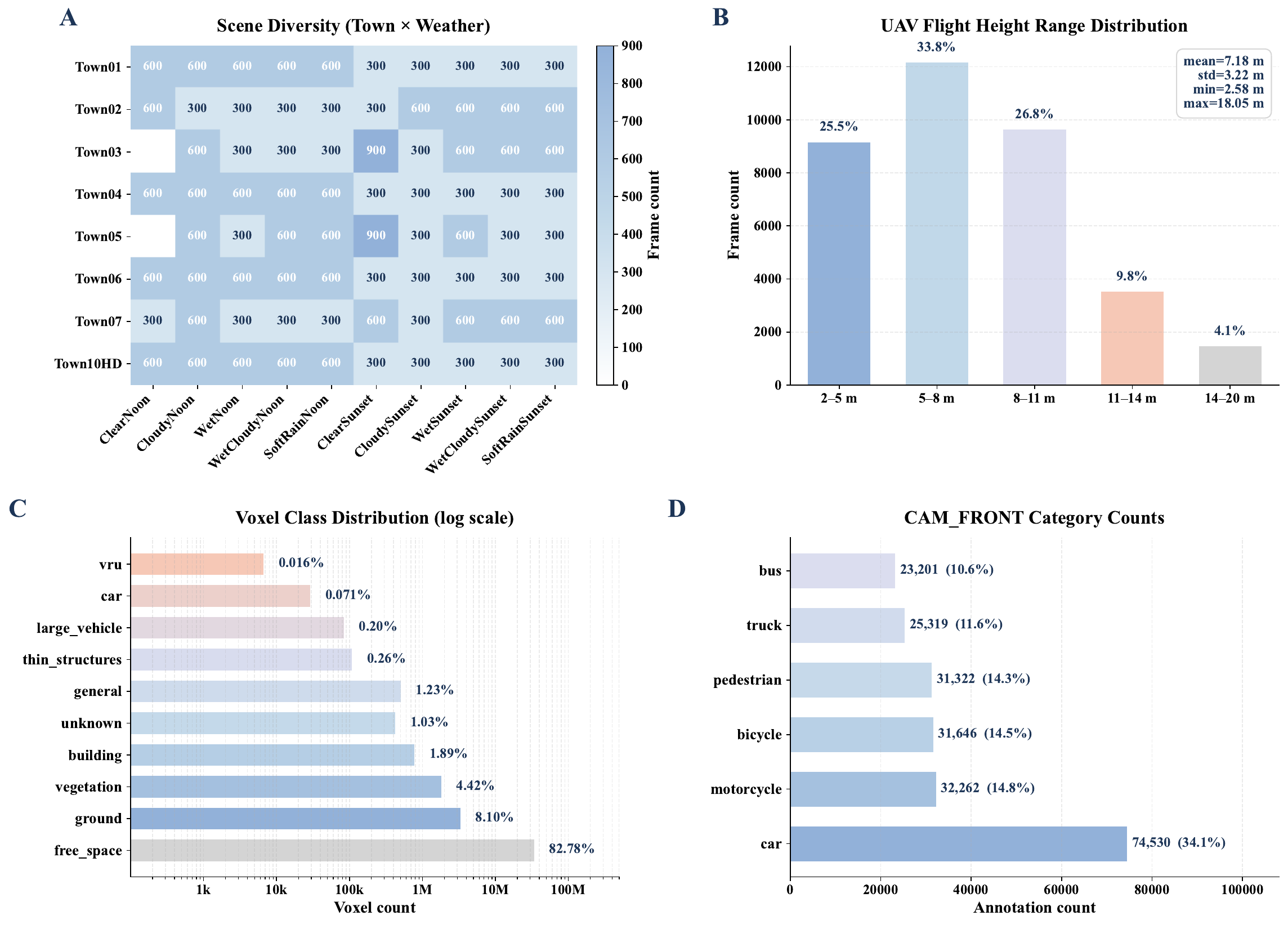}
    \caption{
    \textbf{SkyShield dataset statistics reveal the voxel--safety imbalance of low-altitude UAV occupancy.}
    (A) Scene-weather matrix over 8 towns and 10 illumination-weather conditions, showing diverse environmental coverage.
    (B) UAV flight-height distribution. Most samples lie in the low-altitude envelope, with mean height 7.18m and 86.1\% of frames below 11m.
    (C) Semantic voxel distribution on a logarithmic scale. Free space and large background regions dominate the 3D volume, while safety-critical classes such as thin structures and VRUs are volumetrically rare.
    (D) CAM\_FRONT instance counts. Dynamic objects are frequent in the front view, yet many of them occupy only a negligible fraction of the 3D voxel space.
    }
    \label{fig:app_dataset_statistics}
\end{figure}

\Cref{fig:app_dataset_statistics} reveals the central statistical challenge of low-altitude UAV occupancy. The scene-weather matrix and height distribution establish the physical regime of the benchmark: SkyShield concentrates on city-level flight inside a low-altitude envelope, with a mean height of 7.18m and 86.1\% of frames below 11m. The orthogonal town-weather coverage further reduces the risk that a model overfits to a narrow illumination or scene condition. Within this regime, however, the dataset exposes a sharp mismatch between 2D instance richness and 3D voxel occupancy. CAM\_FRONT contains frequent dynamic actors, including 74,530 cars, 32,262 motorcycles, 31,646 bicycles, 31,322 pedestrians, 25,319 trucks, and 23,201 buses. In stark contrast, when the scene is represented as dense semantic occupancy, free space alone accounts for 82.78\% of voxels, while thin structures, cars, and VRUs occupy only 0.26\%, 0.071\%, and 0.016\%, respectively. This instance--voxel discrepancy is not an incidental dataset artifact; it is a physical property of low-altitude urban flight. Objects that constrain safe navigation can be frequent in the image stream but volumetrically negligible in 3D space. Consequently, a model optimized only for average voxel accuracy can appear strong while still missing the structures that define near-field safety. This voxel--safety imbalance motivates safety-aware evaluation through KAR-mIoU and safety-prior supervision for sparse collision-critical categories.

SkyShield synthesizes UAV motion within a physics-grounded flight envelope. Vertical motion obeys bounded ascent and descent rates:
\begin{equation}
    \Delta h_t
    =
    \operatorname{clip}
    \left(
        h_t^* - h_{t-1},
        -v_{\mathrm{desc}}\Delta t,
        v_{\mathrm{climb}}\Delta t
    \right),
\end{equation}
where $h_t$ is the current altitude, $h_t^*$ is the target altitude, and $\Delta t$ is the simulation interval. Local ray casting estimates nearby obstacle heights $h_t^{\mathrm{obs}}$, and the target altitude is adjusted with a speed-dependent clearance margin:
\begin{equation}
    h_t^*
    =
    \max
    \left(
        h_t^{\mathrm{ref}},
        h_t^{\mathrm{obs}} + d_{\min} + \lambda_h \|\mathbf{v}_t\|_2
    \right).
\end{equation}
Pitch and roll perturbations are generated by an Ornstein--Uhlenbeck process,
\begin{equation}
    d\eta_t
    =
    \kappa(\mu-\eta_t)dt + \sigma dW_t,
\end{equation}
which produces temporally correlated attitude changes rather than independent frame-wise jitter. Together, these constraints keep the synthetic camera trajectory inside a physically plausible low-altitude flight envelope.

For each frame, SkyShield stores the dynamic UAV body pose, fixed front-camera calibration, UAV motion states, semantic observations, actor states, and visibility information.
The full coordinate convention and projection chain used for camera pose derivation, voxel projection, frustum filtering, and visibility validation are provided in Appendix~\ref{app:coordinate_convention}.

\begin{figure}[h]
    \centering
    \includegraphics[width=\textwidth]{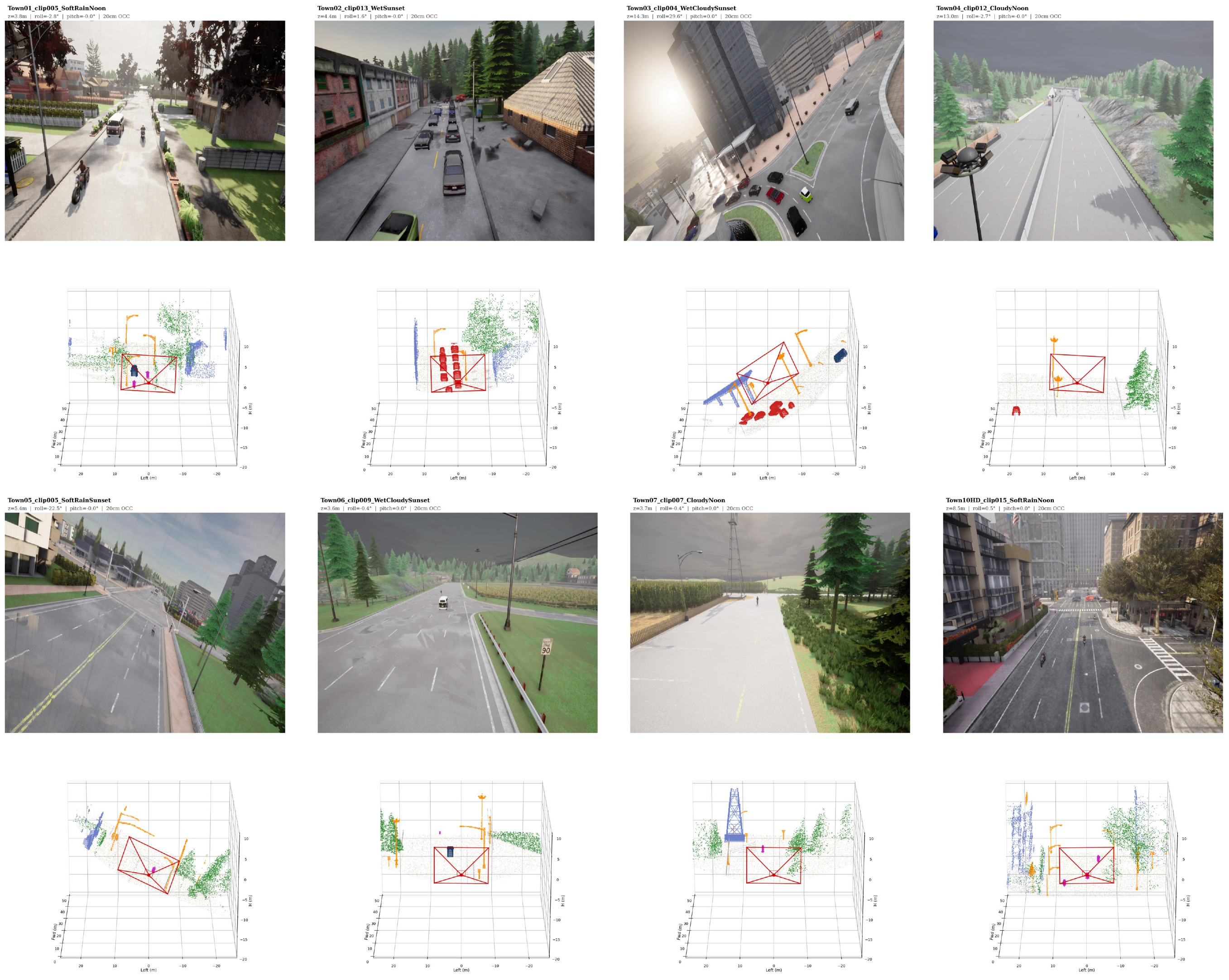}
    \caption{
    \textbf{Kinematic--geometric coupling between UAV attitude, front-view RGB observations, and 3D occupancy labels.}
    Each group pairs a front-view RGB frame with the corresponding UAV-centered semantic occupancy visualization and front-camera frustum. Across diverse environmental and attitude conditions, the red frustum changes with the instantaneous camera pose rather than assuming a static ground-parallel sensor rig. The examples illustrate that low-altitude UAV perception is governed by frame-wise dynamic camera geometry: a tilted image horizon corresponds to a changed projection between the front camera and the UAV-centered occupancy frame.
    }
    \label{fig:app_kinematic_geometric_coupling}
\end{figure}

\Cref{fig:app_kinematic_geometric_coupling} makes the dynamic-extrinsic formulation concrete. The paired RGB and occupancy views show that UAV attitude is not a cosmetic image perturbation; it dictates the physical mapping from front-view pixels to the reachable 3D safety space. In high-roll or high-pitch cases, the RGB horizon becomes strongly tilted, and the corresponding front-camera frustum is spatially warped in the UAV-centered coordinate system. Consequently, monocular occupancy for UAVs cannot rely on a static-rig or ground-parallel projection assumption. The same figure also illustrates why privileged geometric annotation is necessary: even under visual degradation from rain, sunset lighting, vegetation clutter, or severe attitude changes, the occupancy label remains anchored by the frame-wise pose, frame-wise dynamic camera geometry, and semantic geometry. This kinematic--geometric coupling is the physical basis of SkyShield: the benchmark does not only collect diverse images, but enforces a consistent pixel-to-voxel mapping under dynamic low-altitude flight.

\subsection{Coordinate Convention and Projection Chain}
\label{app:coordinate_convention}

We use $\mathbf{T}^{\mathcal{A}\leftarrow\mathcal{B}}\in SE(3)$ to denote a rigid transformation that maps a homogeneous point from frame $\mathcal{B}$ to frame $\mathcal{A}$:
\begin{equation}
    \bar{\mathbf{x}}^{\mathcal{A}}
    =
    \mathbf{T}^{\mathcal{A}\leftarrow\mathcal{B}}
    \bar{\mathbf{x}}^{\mathcal{B}},
    \qquad
    \bar{\mathbf{x}}=[\mathbf{x}^{\top},1]^{\top}.
    \label{eq:app_coord_convention}
\end{equation}
We define four coordinate frames: the world frame $\mathcal{W}$, the UAV body frame $\mathcal{B}_t$, the front-camera frame $\mathcal{C}_t$, and the UAV-centered occupancy frame $\mathcal{G}_t$.
The front camera is rigidly mounted on the UAV; therefore the camera-to-body calibration $\mathbf{T}^{\mathcal{B}\leftarrow\mathcal{C}}$ is time-invariant.
The frame-wise dynamic quantity is the UAV body pose in the world:
\begin{equation}
    \mathbf{T}^{\mathcal{W}\leftarrow\mathcal{B}}_t
    =
    \begin{bmatrix}
    \mathbf{R}^{\mathcal{W}\leftarrow\mathcal{B}}_t & \mathbf{p}^{\mathcal{W}}_t \\
    \mathbf{0}^{\top} & 1
    \end{bmatrix},
    \qquad
    \mathbf{R}^{\mathcal{W}\leftarrow\mathcal{B}}_t
    =
    \mathbf{R}(\mathbf{q}^{\mathrm{uav}}_t),
    \label{eq:app_uav_body_pose}
\end{equation}
where $\mathbf{p}^{\mathcal{W}}_t$ is the UAV position in the world frame and $\mathbf{q}^{\mathrm{uav}}_t$ is the UAV orientation quaternion.
The Euler angles $(\phi_t,\theta_t,\psi_t)$, corresponding to roll, pitch, and yaw, are only a parameterization of $\mathbf{q}^{\mathrm{uav}}_t$ and are not treated as independent pose variables.

The frame-wise front-camera pose is derived from the dynamic UAV body pose and the fixed camera calibration:
\begin{equation}
    \mathbf{T}^{\mathcal{W}\leftarrow\mathcal{C}}_t
    =
    \mathbf{T}^{\mathcal{W}\leftarrow\mathcal{B}}_t
    \mathbf{T}^{\mathcal{B}\leftarrow\mathcal{C}},
    \qquad
    \mathbf{T}^{\mathcal{C}\leftarrow\mathcal{W}}_t
    =
    \left(
    \mathbf{T}^{\mathcal{W}\leftarrow\mathcal{C}}_t
    \right)^{-1}.
    \label{eq:app_dynamic_camera_geometry}
\end{equation}
We use ``dynamic camera geometry'' to denote this frame-wise camera pose induced by UAV motion, rather than a time-varying camera-to-body calibration.

The occupancy grid is represented in a UAV-centered gravity-stabilized frame $\mathcal{G}_t$.
Its origin is the UAV center, its horizontal axes follow the UAV yaw direction, and its vertical axis is aligned with gravity:
\begin{equation}
    \mathbf{T}^{\mathcal{W}\leftarrow\mathcal{G}}_t
    =
    \begin{bmatrix}
    \mathbf{R}_z(\psi_t) & \mathbf{p}^{\mathcal{W}}_t \\
    \mathbf{0}^{\top} & 1
    \end{bmatrix},
    \label{eq:app_occupancy_frame}
\end{equation}
where $\psi_t$ is the UAV yaw angle.
The transformation from the occupancy frame to the front-camera frame is
\begin{equation}
    \mathbf{T}^{\mathcal{C}\leftarrow\mathcal{G}}_t
    =
    \mathbf{T}^{\mathcal{C}\leftarrow\mathcal{B}}
    \mathbf{T}^{\mathcal{B}\leftarrow\mathcal{W}}_t
    \mathbf{T}^{\mathcal{W}\leftarrow\mathcal{G}}_t,
    \label{eq:app_occupancy_to_camera}
\end{equation}
where
$\mathbf{T}^{\mathcal{C}\leftarrow\mathcal{B}}=(\mathbf{T}^{\mathcal{B}\leftarrow\mathcal{C}})^{-1}$
and
$\mathbf{T}^{\mathcal{B}\leftarrow\mathcal{W}}_t=(\mathbf{T}^{\mathcal{W}\leftarrow\mathcal{B}}_t)^{-1}$.
Roll and pitch enter this projection chain exactly once through $\mathbf{T}^{\mathcal{W}\leftarrow\mathcal{B}}_t$.

For a 3D reference point $\mathbf{x}^{\mathcal{G}}_{p,j}$ in the occupancy frame, its image location is computed as
\begin{equation}
    \mathbf{u}_{p,j,t}
    =
    \pi
    \left(
    \mathbf{K}_t
    \begin{bmatrix}
    \mathbf{I}_3 & \mathbf{0}
    \end{bmatrix}
    \mathbf{T}^{\mathcal{C}\leftarrow\mathcal{G}}_t
    \bar{\mathbf{x}}^{\mathcal{G}}_{p,j}
    \right),
    \label{eq:app_attitude_aware_projection}
\end{equation}
where $\mathbf{K}_t\in\mathbb{R}^{3\times3}$ is the front-camera intrinsic matrix, $\bar{\mathbf{x}}^{\mathcal{G}}_{p,j}$ is the homogeneous voxel reference point, $\mathbf{u}_{p,j,t}$ is its projected image coordinate, and $\pi(\cdot)$ denotes perspective normalization.
This single projection chain is used for frustum filtering, visibility validation, and attitude-aware feature lifting.
No additional roll/pitch correction is applied outside this transformation.

\subsection{Annotation and Occupancy Label Generation}
\label{app:annotation_generation}

All labels are generated through a single coordinate chain with explicit frames:
\begin{equation}
    \mathcal{W}
    \xrightarrow{\;\mathbf{T}^{\mathcal{G}\leftarrow\mathcal{W}}_t\;}
    \mathcal{G}_t
    \xrightarrow{\;\mathbf{T}^{\mathcal{C}\leftarrow\mathcal{G}}_t\;}
    \mathcal{C}_t .
    \label{eq:annotation_chain}
\end{equation}
World-space semantic geometry is first transformed into the UAV-centered occupancy frame $\mathcal{G}_t$ for voxelization.
The resulting voxel centers are then projected into the front-camera frame $\mathcal{C}_t$ for frustum filtering and visibility validation.
Each frame stores UAV pose, fixed front-camera calibration, derived dynamic camera geometry, semantic observations, actor states, and visibility information.
This single-chain design avoids mixing simulator, UAV-body, occupancy-grid, and camera coordinate systems, and ensures that UAV attitude affects annotation exactly once through $\mathbf{T}^{\mathcal{W}\leftarrow\mathcal{B}}_t$.

SkyShield uses two LiDAR streams mounted above and below the UAV only for label construction. The upper and lower sensors reduce blind spots around overhangs, facades, vegetation, poles, and nearby clutter. These annotation signals are never exposed to the model input; they improve supervision quality while preserving the monocular learning setting. For dynamic actors, instance masks are decoded to obtain pixel-level actor IDs. For actor $a$ in the front camera, visible support is measured as
\begin{equation}
    V_{a,t}^{\mathrm{front}}
    =
    \sum_{(u,v)}
    \mathbb{I}
    \left[
        \mathrm{id}_t(u,v)=a
    \right],
\end{equation}
where $(u,v)$ indexes image pixels and $\mathrm{id}_t(u,v)$ is the decoded actor ID at frame $t$. This continuous visibility signal is more informative than a binary in-frustum flag. Combined with depth-consistency checks, it suppresses phantom annotations for objects that are geometrically projected into the frustum but visually occluded.

Naive multi-frame voxelization creates two artifacts: moving actors leave temporal ghosts, and partially observed objects produce hollow geometry. SkyShield therefore separates static infrastructure and dynamic actors before voxelization. Static geometry is fused in the world frame and cropped back to the current UAV frame. Dynamic actors are accumulated in object-centric coordinates, which reduces view-dependent sparsity and motion-induced artifacts. This completion step is deliberately conservative: it is used to improve voxel closure, not to claim high-fidelity mesh reconstruction.

The fused scene is voxelized in the UAV front-frustum region. Occupied voxels receive semantic labels by voting over point-level semantics. Observed free space is estimated by ray casting from the front camera and annotation sensors, then filtered by the current front-camera frustum. The final label is a sparse semantic occupancy set
\begin{equation}
    O_t^{\mathrm{front}}
    =
    \{(\mathbf{v}_i,y_i)\},
\end{equation}
where $\mathbf{v}_i$ is a voxel coordinate and $y_i$ is either a semantic class label or the observed free-space label. Unobserved regions are ignored rather than treated as calibrated probabilistic unknown space.

\subsection{KAR-mIoU Implementation}
\label{app:kar_miou_details}

KAR-mIoU is implemented as an asymmetric risk-aware evaluation metric. Its design is intentionally conservative: true positives are not assigned extra reward, while false positives and false negatives are penalized according to their temporal proximity to the UAV. This prevents the score from being inflated by easy predictions in safe or empty regions, and instead focuses evaluation on mistakes that occur within the UAV's short-horizon reachable space.

For voxel $i$, let $\mathbf{x}_i$ denote its center in the UAV-centered frame and $\mathbf{v}_t$ denote the UAV velocity at frame $t$. We first define a clipped evaluation speed
\begin{equation}
    \bar{v}_t
    =
    \max
    \left(
        \|\mathbf{v}_t\|_2,
        v_{\min}
    \right),
\end{equation}
where $v_{\min}$ avoids unrealistically small TTC values caused by near-hovering frames and keeps the safety penalty active within a meaningful local region. The omnidirectional worst-case time-to-collision is then computed as
\begin{equation}
    \mathrm{TTC}^{\mathrm{owc}}_i
    =
    \frac{
        \|\mathbf{x}_i\|_2
    }{
        \bar{v}_t
    } .
\end{equation}
The omnidirectional form avoids assuming that risk only lies along the optical axis. It also accounts for lateral drift, hovering instability, and attitude-induced displacement, which are common in low-altitude UAV flight. Given a safety horizon $T_h$, the reachable safety bubble is
\begin{equation}
    \mathcal{B}_t
    =
    \left\{
        i
        \mid
        \mathrm{TTC}^{\mathrm{owc}}_i
        \leq
        T_h
    \right\}
    =
    \left\{
        i
        \mid
        \|\mathbf{x}_i\|_2
        \leq
        \bar{v}_t T_h
    \right\}.
\end{equation}
Under the minimum evaluation speed, this corresponds to a $25$m effective safety radius.

The penalty for voxel $i$ is defined with an exponential TTC decay:
\begin{equation}
    p_i
    =
    1
    +
    \mathbb{I}
    \left[
        \mathrm{TTC}^{\mathrm{owc}}_i
        \leq
        T_h
    \right]
    \gamma
    \exp
    \left(
        -\lambda
        \mathrm{TTC}^{\mathrm{owc}}_i
    \right),
\end{equation}
where $\gamma$ controls the maximum penalty scale and $\lambda$ controls temporal decay. In our experiments, we use
\begin{equation}
    T_h = 5.0\mathrm{s},
    \qquad
    \lambda = 0.5,
    \qquad
    \gamma = 2.0,
    \qquad
    v_{\min} = 5.0\mathrm{m/s}.
\end{equation}
Thus, the penalty is
\begin{equation}
    p_i
    =
    1
    +
    2.0
    \exp
    \left(
        -0.5
        \mathrm{TTC}^{\mathrm{owc}}_i
    \right),
    \quad
    \mathrm{if}
    \quad
    \mathrm{TTC}^{\mathrm{owc}}_i
    \leq
    5.0\mathrm{s},
\end{equation}
and $p_i=1$ outside the safety horizon. The maximum penalty is therefore $3\times$ at zero TTC and decays with temporal distance.

For semantic class $c$, KAR-IoU is computed as
\begin{equation}
    \mathrm{KAR\text{-}IoU}_c
    =
    \frac{
        \mathrm{TP}_c
    }{
        \mathrm{TP}_c
        +
        \sum_{i\in\mathrm{FP}_c} p_i
        +
        \sum_{i\in\mathrm{FN}_c} p_i
    } .
\end{equation}
Only false positives and false negatives are risk-weighted. Correct predictions are deliberately left unweighted, so the metric does not reward the model for easy near-field true positives; it only increases the cost of safety-relevant mistakes.

Unknown voxels are excluded from evaluation. Free space is reported separately but excluded from the KAR-mIoU mean because it dominates the voxel distribution. False positives and false negatives involving occupied classes remain in the class-wise denominators, so predicting an occupied object as free space is still penalized as a false negative for the corresponding occupied class. The final KAR-mIoU is obtained by averaging KAR-IoU over evaluated occupied semantic classes. The complete parameter configuration for the KAR-mIoU evaluation is summarized in \Cref{tab:app_kar_config}.

\begin{table}[h]
\centering
\small
\caption{\textbf{KAR-mIoU evaluation configuration.} KAR-mIoU uses only kinematic risk penalties during evaluation; class weights are not used in the metric.}
\label{tab:app_kar_config}
\begin{tabular}{lcl}
\toprule
Parameter & Value & Role \\
\midrule
Penalty function & Exponential & TTC-decayed FP/FN penalty \\
$T_h$ & 5.0s & Short-horizon safety window \\
$\lambda$ & 0.5 & TTC penalty decay \\
$\gamma$ & 2.0 & Maximum additional penalty scale \\
$v_{\min}$ & 5.0m/s & Minimum evaluation speed for TTC \\
Effective radius & 25m & Local safety region under $v_{\min}$ \\
Exclude free space & True & Avoid domination by empty-space voxels \\
Exclude unknown & True & Ignore unreliable unknown regions \\
Penalized mode & True & Penalize FP/FN only; do not up-weight TP \\
\bottomrule
\end{tabular}
\end{table}

Thus, KAR-mIoU asks a stricter safety question than standard mIoU: when occupancy prediction fails, does the failure occur in the region where the UAV has the least time to recover?

\subsection{SkyOcc Implementation Details}
\label{app:skyocc_details}

SkyOcc is implemented as a geometry-aware monocular UAV-OCC reference baseline.
For all comparisons, the baseline uses the same front-view RGB input, temporal queue, image backbone, FPN neck, temporal pillar encoder, and occupancy decoder as SkyOcc, but does not include attitude-aware projection or safety-prior optimization.
We then progressively add attitude-aware projection, static safety-prior class weights, and Lov\'asz-Softmax loss.
The final configuration is denoted as \textbf{SkyOcc}.

The implementation uses a ResNet50 image backbone, an FPN neck with four feature levels, a spatiotemporal pillar encoder, and a semantic voxel decoder. The temporal queue length is 4. The voxel grid size is $100\times100\times60$ with 0.5m resolution, and the point-cloud range is $[0,-25,-20,50,25,10]$ meters. Historical pillar features are aligned to the current UAV-centered coordinate system before temporal aggregation, while attitude-aware projection is used before front-view feature lifting. This keeps the architecture close to a standard BEVFormer occupancy baseline while isolating the effect of geometry correction under dynamic UAV attitude.

For optimization, SkyOcc uses class-balanced cross-entropy with safety-prior weights and Lov\'asz-Softmax loss:
\begin{equation}
    \mathcal{L}_{\mathrm{occ}}
    =
    \frac{1}{|\Omega|}
    \sum_{i\in\Omega}
    \alpha_{y_i}
    \operatorname{CE}
    \left(
        \hat{\mathbf{y}}_i,
        y_i
    \right)
    +
    \lambda_{\mathrm{lov}}
    \mathcal{L}_{\mathrm{lov}}
    \left(
        \hat{\mathbf{Y}}_t,
        \mathbf{Y}_t
    \right),
\end{equation}
where $\Omega=\{i\mid y_i\neq c_{\mathrm{ignore}}\}$ is the supervised voxel set, $\alpha_{y_i}$ is a safety-prior class weight, and $\mathcal{L}_{\mathrm{lov}}$ is an IoU-oriented surrogate. We set $\lambda_{\mathrm{lov}}=1$ in all experiments. The class weights are selected from training-set occupancy statistics: we first compute the frequency of each semantic occupancy class, then assign larger weights to categories that are both rare in voxel space and safety-critical for low-altitude flight, such as thin structures, vehicles, and VRUs.

This design shifts learning pressure from dominant regions such as free space, ground, and buildings toward sparse collision-critical categories. It preserves the monocular benchmark protocol while making the baseline better aligned with the safety bottlenecks revealed by the voxel distribution of low-altitude UAV scenes.

All models are trained on 8 NVIDIA A800 GPUs with per-GPU batch size 8, giving a total batch size of 64. Each model is trained for 24 epochs using AdamW with initial learning rate $2\times10^{-4}$, weight decay $1\times10^{-2}$, and cosine annealing learning-rate scheduling.

\subsection{Complementary Experimental Protocol and Additional Results}
\label{app:additional_results}

Since the main paper reports the primary mIoU/KAR-mIoU comparison and the main ablation, we provide complementary views here to make the evaluation path explicit. We focus on the multi-frame progression, long-tail safety-category behavior, and step-wise component effects. These results show not only how the final score improves, but also where the improvement is produced in the occupied space.

\Cref{tab:app_test_summary} summarizes the multi-frame progression. Starting from a temporal monocular occupancy baseline, we progressively add attitude-aware projection, static safety-prior class weights, and Lov\'asz-Softmax loss. The final configuration improves the multi-frame baseline by +3.83 mIoU and +4.06 KAR-mIoU. The mIoU--KAR-mIoU gap remains close to 10 points across variants, suggesting that these components consistently improve occupied prediction quality, while near-field risk concentration remains a distinct challenge that is not fully solved by global reconstruction losses alone.

\begin{table}[h]
\centering
\small
\caption{\textbf{Complementary multi-frame progression on the test split.} $\Delta$ values are relative to the multi-frame baseline. All numbers are percentages.}
\label{tab:app_test_summary}
\resizebox{\linewidth}{!}{
\begin{tabular}{lccccc}
\toprule
Configuration & \textbf{mIoU} & \textbf{KAR-mIoU} & $\Delta$ \textbf{mIoU} & $\Delta$ \textbf{KAR} & \textbf{Gap} \\
\midrule
Multi-frame baseline & 25.08 & 15.04 & 0.00 & 0.00 & 10.04 \\
+ Attitude-aware projection & 26.56 & 16.59 & +1.48 & +1.55 & 9.97 \\
+ Safety-prior class weights & 28.04 & 18.17 & +2.96 & +3.13 & 9.86 \\
+ Lov\'asz-Softmax & \textbf{28.91} & \textbf{19.10} & \textbf{+3.83} & \textbf{+4.06} & \textbf{9.81} \\
\bottomrule
\end{tabular}
}
\end{table}

The strongest improvements appear in long-tail safety categories. \Cref{tab:app_long_tail} isolates thin structures, large vehicles, cars, and VRUs, which occupy few voxels but often define the boundary of safe low-altitude flight. Compared with the multi-frame baseline, the final model improves these categories by +7.23, +4.66, +10.05, and +9.14 points, respectively. This shows that SkyShield does not merely improve average scene reconstruction; it reallocates representational capacity toward sparse occupied structures that are easy to miss but costly to ignore.

\begin{table}[h]
\centering
\small
\caption{\textbf{Long-tail safety structures on the test split.} Gains are relative to the multi-frame baseline. All numbers are standard IoU percentages.}
\label{tab:app_long_tail}
\resizebox{\linewidth}{!}{
\begin{tabular}{lccccc}
\toprule
Configuration & Thin Struct. & Large Veh. & Car & VRU & Mean \\
\midrule
Multi-frame baseline & 5.57 & 10.16 & 13.19 & 2.95 & 7.97 \\
+ Attitude-aware projection & 7.17 & 11.39 & 15.32 & 9.54 & 10.86 \\
+ Safety-prior class weights & 11.85 & 14.66 & 20.90 & 11.12 & 14.63 \\
\textbf{+ Lov\'asz-Softmax} & \textbf{12.80} & \textbf{14.82} & \textbf{23.24} & \textbf{12.09} & \textbf{15.74} \\
\midrule
Gain of full model & +7.23 & +4.66 & +10.05 & +9.14 & +7.77 \\
\bottomrule
\end{tabular}
}
\end{table}

\Cref{tab:app_incremental_effects} further decomposes the contribution of each added component. Attitude-aware projection provides the largest immediate gain on VRU, confirming that measured UAV pose should enter the image-to-3D projection chain explicitly. Safety-prior class weights then produce the strongest gains on thin structures, large vehicles, and cars, showing that long-tail hazards require direct optimization pressure. Lov\'asz-Softmax provides the final IoU-oriented refinement, improving both the global score and the critical-category average without changing the monocular input protocol.

\begin{table}[h]
\centering
\small
\caption{\textbf{Incremental component effects on the test split.} Values are percentage-point changes from the immediately previous multi-frame configuration.}
\label{tab:app_incremental_effects}
\resizebox{\linewidth}{!}{
\begin{tabular}{lcccccc}
\toprule
Added component & $\Delta$\textbf{mIoU} & $\Delta$\textbf{KAR} & $\Delta$Thin Struct. & $\Delta$Large Veh. & $\Delta$Car & $\Delta$VRU \\
\midrule
Attitude-aware projection & +1.48 & +1.55 & +1.60 & +1.23 & +2.13 & +6.59 \\
Safety-prior class weights & +1.48 & +1.58 & +4.68 & +3.27 & +5.58 & +1.58 \\
Lov\'asz-Softmax & +0.87 & +0.93 & +0.95 & +0.16 & +2.34 & +0.97 \\
\bottomrule
\end{tabular}
}
\end{table}

\begin{figure}[h]
    \centering
    \includegraphics[width=\textwidth]{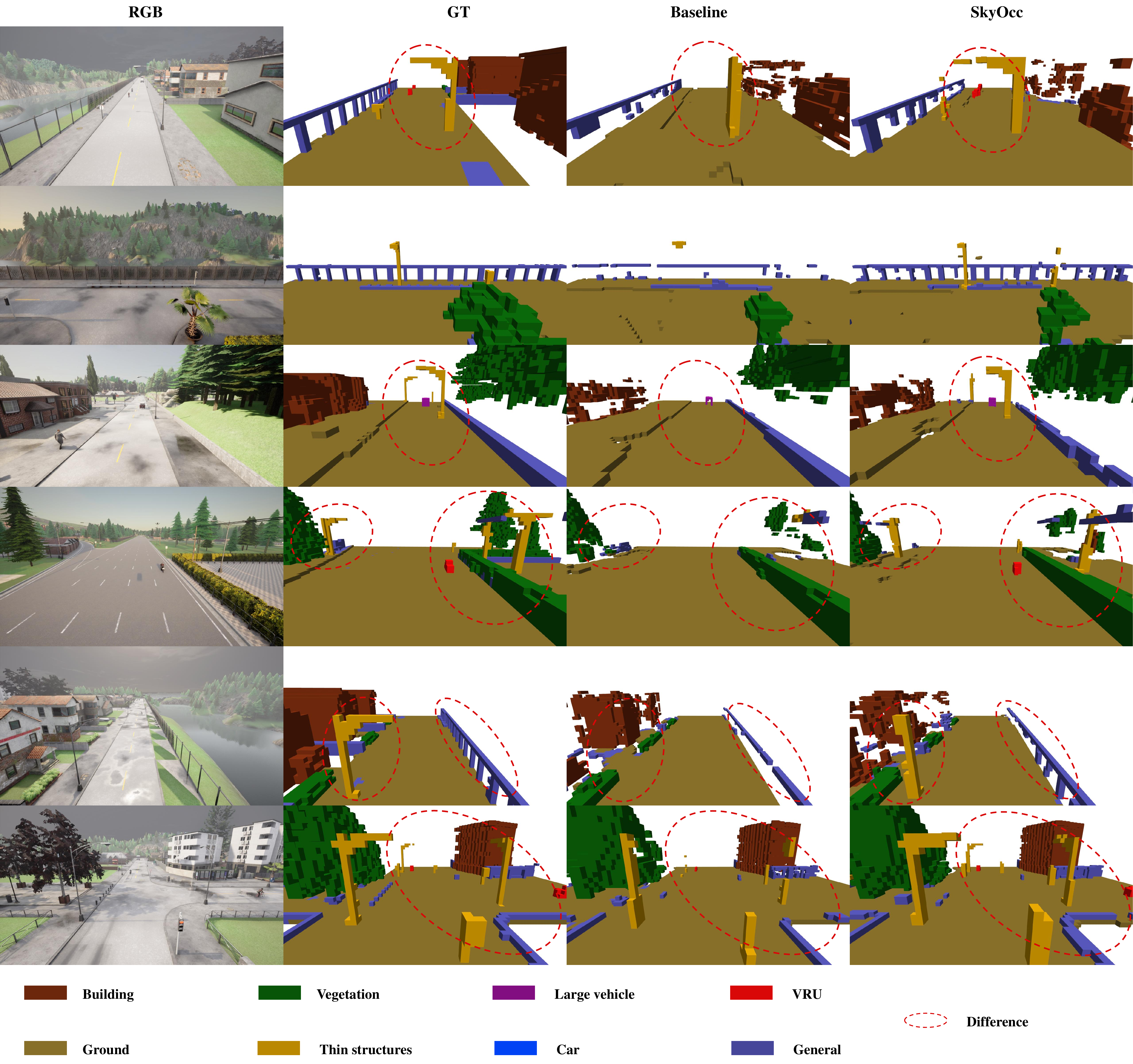}
    \caption{\textbf{Additional qualitative results.} Each group shows the front-view RGB, ground truth, baseline prediction and SkyOcc prediction.}
    \label{fig:app_qualitative}
\end{figure}

\subsection{Additional Qualitative Results and Limitations}
\label{app:qualitative_limitations}

To further examine the behavior of SkyOcc under diverse low-altitude flight conditions, we provide additional qualitative results in \Cref{fig:app_qualitative}. Compared with the representative examples in the main paper, these supplementary cases cover a wider spectrum of challenging urban UAV scenarios, including dense vegetation occlusion, cluttered streets, nearby vehicles, vulnerable road users (VRUs), and sparse safety-critical structures such as poles, traffic signs, wires, fences, and tree branches. These examples reflect the central difficulty of UAV-centric semantic occupancy prediction: many obstacles that are crucial for safe flight occupy only a small number of pixels in the monocular image and an even smaller fraction of voxels in 3D space, yet they may directly determine whether the forward region is traversable.

The qualitative results are consistent with the empirical trends observed in the quantitative experiments. Attitude-aware projection improves the geometric consistency between the front-view image and the UAV-centered occupancy volume, especially under frame-wise changes in pitch, roll, altitude, and camera pose. This reduces projection drift and leads to more stable reconstruction of the forward reachable space. In addition, safety-prior optimization encourages the model to preserve sparse but safety-critical structures that are easily suppressed by dominant classes such as ground, vegetation, buildings, and free space. As a result, SkyOcc produces occupancy predictions that are not only more accurate in an average voxel-wise sense, but also more aligned with the safety requirements of low-altitude autonomous flight.

Despite these encouraging results, SkyShield remains a simulation-based benchmark. Although the CARLA-based generation pipeline enables scalable data collection with diverse towns, weather conditions, trajectories, dynamic actors, and controllable annotation signals, it cannot fully reproduce all real-world sensing effects encountered by physical UAV platforms. Factors such as motion blur, rolling-shutter distortion, camera noise, lens contamination, illumination instability, severe weather degradation, vibration, and rotor-induced airflow may introduce additional challenges when transferring models from simulation to real-world deployment. Bridging this simulation-to-real gap is therefore an important direction for future work, and we hope SkyShield can serve as a controlled starting point for studying robust UAV occupancy learning before extending the task to real-world flight data.

More broadly, SkyOcc is intentionally designed as a transparent, lightweight, and reproducible baseline rather than a final solution for aerial autonomy. Its purpose is to expose the unique challenges introduced by UAV-centered semantic occupancy prediction: dynamic 6-DoF observation, monocular depth ambiguity, sparse safety-critical geometry, severe occlusion, and the need to reason about the 3D space that the vehicle is about to enter. By providing a clear task definition, a dedicated benchmark, safety-aware evaluation, and an interpretable reference model, SkyShield aims to establish a common foundation for the community to study low-altitude spatial perception in a more systematic way.

We believe this task has implications beyond occupancy prediction itself. Reliable understanding of occupied, free, unknown, and safety-critical space is a prerequisite for the next generation of embodied aerial intelligence. As UAVs increasingly interact with high-level human instructions, language-conditioned goals, and open-world environments, dense 3D spatial perception will become a core interface between perception, planning, and action. Future vision-language navigation (VLN), vision-language-action (VLA), and world-model-based UAV systems will require not only recognizing objects in images, but also understanding whether the space ahead can be safely entered, avoided, or used for long-horizon decision-making. In this sense, SkyShield takes an initial step toward connecting low-altitude UAV perception with downstream embodied autonomy.

We hope that SkyShield will encourage broader participation in this emerging direction. By defining front-view monocular semantic occupancy prediction for low-altitude UAV flight, we aim to move the field from image-level aerial recognition toward safety-aware 3D space understanding. Future research may build upon this benchmark to explore stronger temporal reasoning, uncertainty-aware prediction, multi-modal sensor fusion, sim-to-real adaptation, world-model integration, and closed-loop perception-planning evaluation. Ultimately, improving low-altitude UAV spatial perception will provide a critical foundation for safer and more capable aerial agents operating in complex human-scale environments.

\end{document}